\documentclass[gmd, manuscript]{copernicus}
\nolinenumbers

\begin{document}

\title{Convolutional conditional neural processes for local climate downscaling}

% \Author[affil]{given_name}{surname}

\Author[1]{Anna}{Vaughan}
\Author[1]{Will}{Tebbutt}
\Author[2,3]{J. Scott}{Hosking}
\Author[1]{Richard E.}{Turner}

\affil[1]{University of Cambridge, Cambridge, UK}
\affil[2]{British Antarctic Survey, Cambridge, UK}
\affil[3]{The Alan Turing Institute, UK}

\correspondence{Anna Vaughan (av555@cam.ac.uk)}

\runningtitle{*}

\runningauthor{*}

\received{}
\pubdiscuss{} %% only important for two-stage journals
\revised{}
\accepted{}
\published{}

%% These dates will be inserted by Copernicus Publications during the typesetting process.

\firstpage{1}

\maketitle

\begin{abstract}

A new model is presented for multisite statistical downscaling of temperature and precipitation using convolutional conditional neural processes (convCNPs). ConvCNPs are a recently developed class of models that allow deep learning techniques to be applied to off-the-grid spatio-temporal data. This model has a substantial advantage over existing downscaling methods in that the trained model can be used to generate multisite predictions at an arbitrary set of  locations, regardless of the availability of training data. The convCNP model is shown to outperform an ensemble of existing downscaling techniques over Europe for both temperature and precipitation taken from the VALUE intercomparison project. The model also outperforms an approach that uses Gaussian processes to interpolate single-site downscaling models at unseen locations. Importantly, substantial improvement is seen in the representation of extreme precipitation events. These results indicate that the convCNP is a robust downscaling model suitable for generating localised projections for use in climate impact studies, and motivates further research into applications of deep learning techniques in statistical downscaling.

\end{abstract}

\introduction  %% \introduction[modified heading if necessary]
Statistical downscaling methods are vital tools in translating global and regional climate model output to actionable guidance for climate impact studies. General circulation models (GCMs) and regional climate models (RCMs) are used to provide projections of future climate scenarios, however coarse resolution and systematic biases result in unrealistic behaviour, particularly for extreme events \citep{allen2016climate,maraun2017towards}. In recognition of these limitations, downscaling is routinely performed to correct raw GCM and RCM outputs. This is achieved either by dynamical downscaling, running a nested high-resolution simulation, or statistical methods. Comparisons of statistical and dynamical downscaling suggest that neither group of methods is clearly superior \citep{ayar2016intercomparison,casanueva2016towards}, however in practice computationally cheaper statistical methods are widely used. 

Major classes of statistical downscaling methods are model output statistics (MOS) and perfect prognosis (PP; Maraun et al., 2010)\nocite{maraun2010precipitation}. MOS methods explicitly adjust the simulated distribution of a given variable to the observed distribution, using variations of quantile mapping \citep{teutschbein2012bias, piani2010statistical, cannon2020bias}. Though these methods are widely applied in impact studies, they struggle to downscale extreme values and artificially alter trends \citep{maraun2013bias, maraun2017towards}. In contrast, in PP downscaling, the aim is to learn a transfer function f such that  
\begin{equation}
\hat{y} = f(x, Z)
\end{equation}
Where $\hat{y}$ is the downscaled prediction of a given climate variable whose true value is $y$ at location $x$ and Z is a set of predictors from the climate model \citep{maraun2018statistical}. This is based on the assumption that while sub-grid-scale and parameterised processes are poorly represented in GCMs, the large scale flow is generally better resolved. 

Multiple different models have been trialled for parameterising $f$. Traditional statistical methods used for this purpose include multiple linear regression \citep{gutierrez2013reassessing,hertig2013novel}, generalised linear models \citep{san2017reassessing} and analog techniques \citep{hatfield2015temperature,ayar2016intercomparison}. More recently, there has been considerable interest in applying advances in machine learning to this problem, including relevance vector machines \citep{ghosh2008statistical}, artificial neural networks \citep{sachindra2018statistical}, autoencoders \citep{vandal2019intercomparison}, recurrent neural networks \citep{bhardwaj2018downscaling} and convolutional neural networks \citep{bano2020configuration,hohlein2020comparative}. There has been debate as to whether deep learning methods provide improvement over traditional statistical techniques such as multiple linear regression. \cite{vandal2019intercomparison} found that machine learning approaches offered little improvement over traditional methods in downscaling precipitation. In contrast \cite{bano2020configuration} compared downscaling performance of convolutional neural networks (CNNs) and simple linear models, finding that CNNs improved predictions of precipitation, but did not result in improvements for temperature. 

Limitations remain in these models. In many climate applications it is desirable to make projections that are both (i) consistent over multiple locations and (ii) specific to an arbitrary locality. The problem of multi-site downscaling has been widely studied, with two classes of approaches emerging. Traditional methods take analogues or principal components of the coarse-resolution field as predictors. The spatial dependence is then explicitly modelled for a given set of sites, using observations at those locations to train the model \citep{maraun2018statistical,cannon2008probabilistic,bevacqua2017multivariate,mehrotra2005nonparametric}. More recent work has sought to leverage avances in machine learning, for example CNNs, for feature extraction \citep{bhardwaj2018downscaling,bano2020configuration,hohlein2020comparative}. Each of these classes of methods has their drawbacks. Traditional methods have limited feature selection, however are able to train directly on true observations. In contrast, deep learning techniques such as CNNs allow for sophisticated feature extraction, however are only able to be applied to gridded datasets. Gridding of observations naturally introduces error, especially for areas with complex topography and highly stochastic variables such as precipitation \citep{king2013efficacy}. The second limitation common to existing downscaling models is that predictions can only be made at sites for which training data are available. Creating a projection at an arbitrary location is achieved either through interpolation of model predictions or taking the closest location. 

In this study, we address these challenges by developing a new statistical model for downscaling temperature and precipitation at an arbitrary set of sites. This is achieved using convolutional conditional neural processes, state of the art probabilistic machine learning methods combining ideas from Gaussian Processes (GPs) and deep neural networks. The model combines the advantages of existing classes of multi-site approaches, with feature extraction using a convolutional neural network together with training on off-the-grid data. In contrast to existing methods, the trained model can be used to make coherent local projections at any site regardless of the availability of training observations. 

The specific aims of this study are as follows: 

\begin{enumerate}
	\item Develop a new statistical model for downscaling GCM output capable of training on off-grid data, making predictions at unseen locations and utilising sub-grid-scale topographic information to refine local predictions.  
	\item Compare the performance of the statistical model to existing strong baselines. 
	\item Compare the performance of the statistical model at unseen locations  to existing interpolation methods.
	\item Quantify the impact of including sub-grid-scale topography on model predictions.  
\end{enumerate}  

Section 2 outlines the development of the downscaling model, and presents the experimental setup used to address aims 2-4. Section 3 compares the performance of the statistical model to an ensemble of baselines. Sections 4 and 5 explore model performance at unseen locations and the impact of including local topographic data. Finally, Section 6 presents a discussion of these results and suggestions for further applications.

%%%%%%%%%%%%%%%%%%%%%%%%%%%%%%%%%%%
% DATASETS AND METHODOLOGY
%%%%%%%%%%%%%%%%%%%%%%%%%%%%%%%%%%% 

\section{Datasets and methodology}
We first outline the development of the statistical downscaling model, followed by a description of three validation experiments. 

\subsection{The downscaling model}
Our aim is to approximate the function $f$ in equation 1 to predict the value of a downscaled climate variable $y$ at locations $\textbf{x}$ given a set of coarse-scale predictors $Z$. In order to take the local topography into account, we assume that this function also depends on the local topography at each target point, denoted $\textbf{e}$, i.e
\[\hat{y} = f(\textbf{x},Z,\textbf{e})\]

We take a probabilistic approach to specifying f where we include a noise model, so that 
\[p(y|\textbf{x},Z,\textbf{e}) = p(y|\theta(\textbf{x},Z,\textbf{e}))\]
Deterministic predictions are made from this by using, for example, the predictive mean 
\[\hat{y} = \int yp(y|\textbf{x},Z,\textbf{e})dy\]
In this model $\theta$ is parameterised as   
\[ \theta(\textbf{x},Z,\textbf{e}) = \psi_{MLP}[\phi_{c}(h = CNN(Z), \textbf{x}), \textbf{e}]\]
Here $\theta$ is a vector of parameters of an distribution for the climate variable at prediction locations $\textbf{x}$. This is assumed to be Gaussian for maximum temperature and a Gamma-Bernoulli mixture for precipitation. $\textbf{e}$ is a vector of sub-grid-scale topographic information at each of the prediction locations, $\psi_{MLP}$ is a multi-layer perceptron, $\phi_{c}$ is a function parameterised as a neural network and CNN is a convolutional neural network. Each component of this is described below, with a schematic of the model shown in Figure 1. 

\begin{figure}
\centering
\includegraphics[scale=0.15]{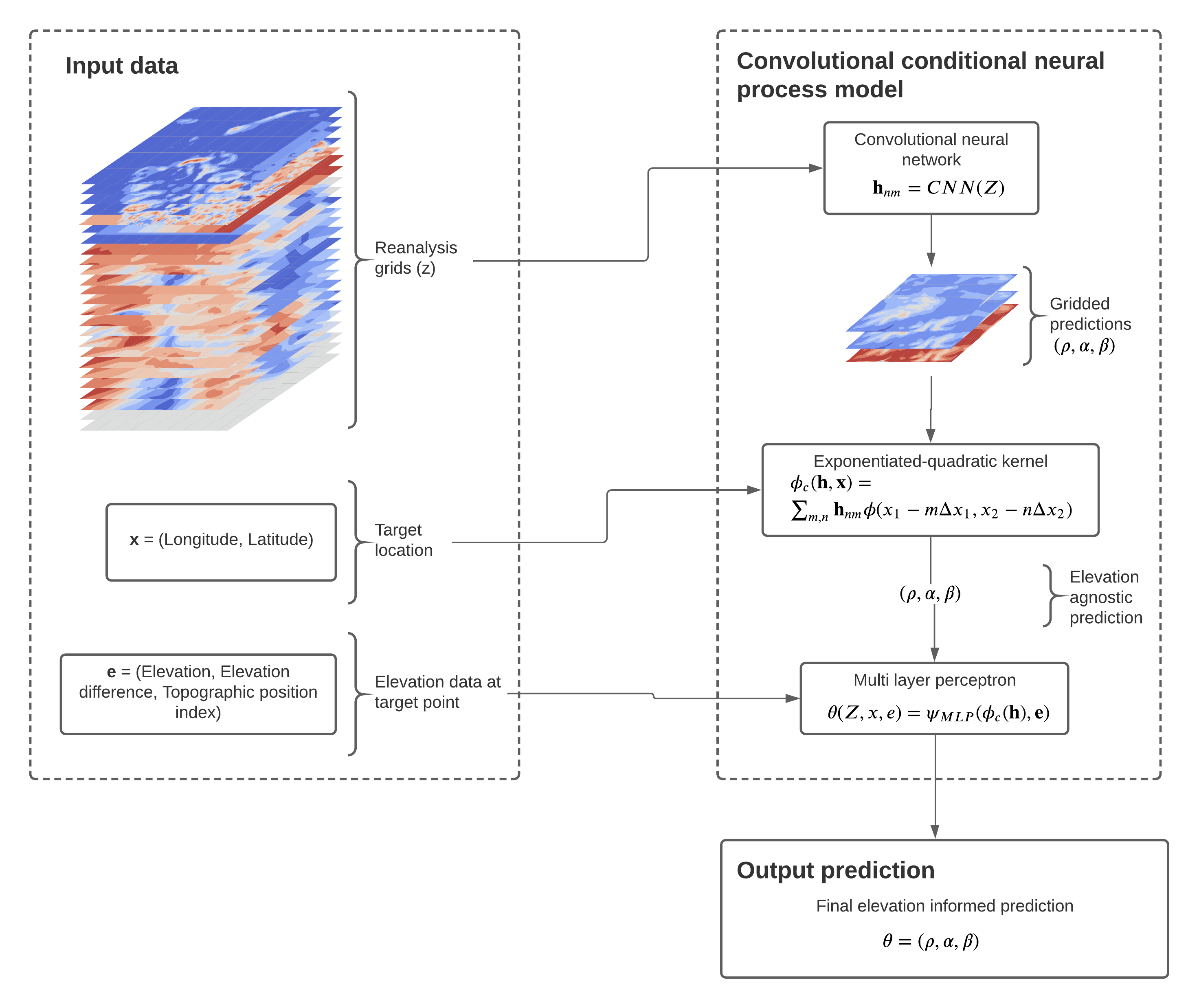}
\caption{Schematic of the convCNP model for downscaling precipitation demonstrating the flow of data in predicting precipitation for a given day at target locations $\textbf{x}$. Gridded coarse resolution data for each predictor is fed into the CNN, producing predictions of $\theta = (\rho, \alpha, \beta)$ at each grid point. These gridded predictions are then transformed to a prediction at the target location using an exponentiated-quadratic kernel. Finally, these elevation agnostic predictions are fed into a multi-layer perceptron together with topographic data $\textbf{e}$ to produce a final prediction of the parameters.}	
\end{figure}

\begin{enumerate}

\item \textbf{Convolutional neural network}
In the first step, gridded reanalysis predictor data $Z$ are fed into the model. These grids are used as input to a convolutional neural network to extract relevant features. This is implemented as a 6-block Resnet architecture \citep{he2016deep} with depthwise separable convolutions \citep{chollet2017xception}. The output from this step is a prediction of the relevant parameters for each variable at each grid point in the predictor set, i.e
\[\textbf{h}_{nm} = CNN(Z)\]
Where $\textbf{h}_{nm}$ is the vector-valued output at latitude $m\Delta x_1$ and longitude $n\Delta x_2$ and $m,n \in \mathbb{Z}^{+}$ with $\Delta x_i$ indicating the grid spacing.

\item \textbf{Translation to off-the-grid predictions}
These gridded predictions are translated to the off-the-grid target locations $\textbf{x}$ using outputs from step 1 as weights for an exponentiated-quadratic (EQ) kernel $\phi$, i.e
\[\phi_{c}(\textbf{h},\textbf{x}) = \sum_{m,n}\textbf{h}_{nm}\phi(x_1-m\Delta x_1, x_2-n\Delta x_2) = \sum_{m,n}\textbf{h}_{nm}e^{-\frac{1}{2l_1^2}(x_1-m\Delta x_1)^2 -\frac{1}{2l_2^2}(x_2-n\Delta x_2)^2}\] 
 This outputs predictions of the relevant distributional parameters at each target location $\theta$. 

\item \textbf{Inclusion of sub-grid scale topography}

By design, the predictions from the previous step only model variation on the scale of the context grid spacing. This elevation agnostic output is post-processed using a multi-layer perceptron (MLP). This takes the parameter predictions from the EQ kernel as input together with a vector of topographic data $\textbf{e}$ at each target location. 
\[ \theta(\textbf{x},Z,\textbf{e}) = \psi_{MLP}(\phi_c(\textbf{h}),\textbf{e})\]
The vector $\textbf{e}$ consists of three measurements at each target point:
\begin{enumerate}
	\item True elevation
	\item Difference between the true and grid-scale elevation. 
	\item Multi-scale topographic position index (mTPI), measuring the topographic prominence of the location, (i.e quantifying whether the point is in a valley or on a ridge). 
\end{enumerate}
This MLP outputs the final prediction of the distributional parameters $\theta$.

\end{enumerate}

Figure 2 shows a concrete example of temperature and precipitation time-series produced using this model by sampling from the output distributions. Maximum temperature is shown for Helgoland, Germany, and precipitation for Madrid, Spain. For both variables the model produces qualitatively realistic time-series. 

\begin{figure}
\begin{center}
\includegraphics[scale=0.40]{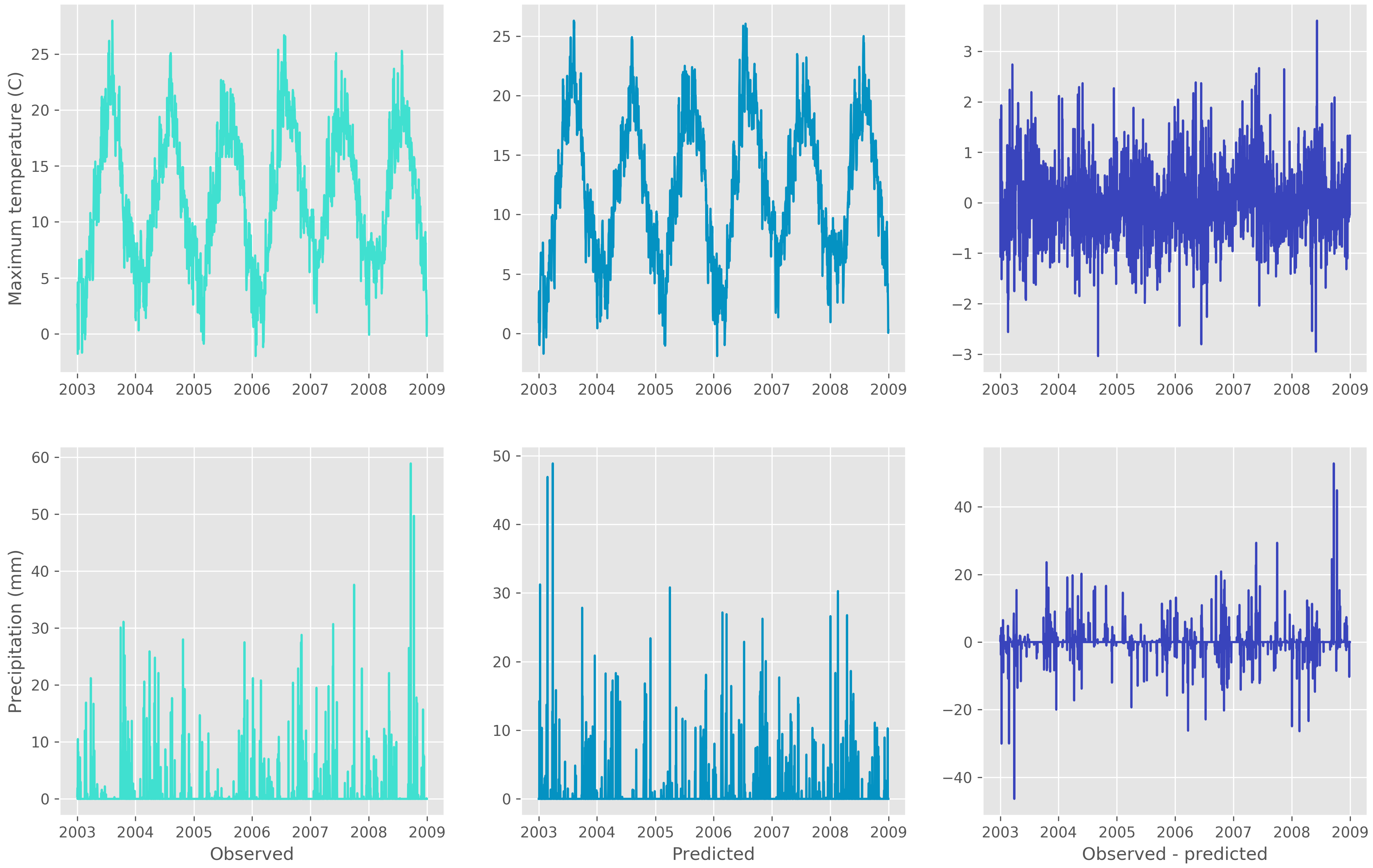}
\caption{Examples of convCNP model predictions compared to observations for (top) maximum temperature in Helgoland, Germany and (bottom) precipitation in Madrid, Spain.}	
\end{center}
\end{figure}

This model is an example of a more general class of models known as convolutional conditional neural processes (ConvCNPs; Gordon et al., 2019). Formally, PP downscaling is an instance of a supervised learning problem to learn the function $f$ in equation 1. Approaches to learning such a function have traditionally been split into two categories: parametric approaches, where $f$ is parameterised by a neural network which is optimised by gradient descent, and Bayesian methods specifying a distribution over $f$. Conditional neural processes (CNPs; Garnelo et al., 2018) combine the advantages of these approaches, using neural networks to parameterise a stochastic process over the output variable, in this case either temperature or precipitation.  Various improvements to the original CNP model have been suggested, for example the inclusion of a global latent variable \citep{garnelo2018neural} and attention to improve the flexibility of the model and prevent under-fitting \citep{kim2019attentive}. In the context of spatial data such as downscaling predictions, a desirable characteristic of predictions are that they are translation-equivariant, that is the model makes identical predictions if data are spatially translated. The convCNP model \citep{gordon2019convolutional} applied here builds this equivariance into the CNP model. Throughout this study, we refer to the model developed in this section as `the convCNP model'.  

\subsubsection{Training}

The convCNP models are trained by minimising the average negative log-likelihood. For temperature, this is given by   
\[ NLL_{temp} = -\frac{1}{N}\sum_{i=1}^{N} log[\mathcal{N}(y_{i}|\mu_{i}(\textbf{x}_{i}, Z, \textbf{e}_i), \sigma_{i}(\textbf{x}_{i}, Z, \textbf{e}_i))\]
where $y_i$ is the observed value, $\mathcal{N}(y_{i};\mu_{i},\sigma_{i})$ denotes a Gaussian distribution over $y$ with mean $\mu_{i}$ and variance $\sigma_{i}^2$. These parameters $\theta(\textbf{x}_i,Z,\textbf{e}_i) = \{\mu_i(\textbf{x}_i,Z,\textbf{e}_i), \sigma_i^2(\textbf{x}_i,Z,\textbf{e}_i)\}$ are generated by the model at each location $\textbf{x}_i$ and use topography $\textbf{e}_i$. N is the total number of target locations. For precipitation, the negative log likelihood is given by
\[NLL_{precip} = -\frac{1}{N}\sum_{i=1}^{N}[r_i(ln(\rho_i(\textbf{x}_{i}, Z, \textbf{e}_i))+ln(\Gamma(y_i |\alpha_i(\textbf{x}_{i}, Z, \textbf{e}_i),\beta_i(\textbf{x}_{i}, Z, \textbf{e}_i)))+(1-r_i)ln(1-\rho_i(\textbf{x}_{i}, Z, \textbf{e}_i))]\]
where $r_i$ is a Bernoulli random variable describing whether precipitation was observed at the $i^{th}$ target location, $y_i$ is the observed precipitation, $\rho_i$ parameterises the predicted Bernoulli distribution and $\Gamma(y_i;\alpha_i,\beta_i)$ is a Gamma distribution with shape parameter $\alpha_i$ and scale parameter $\beta_i$. Here $\theta(\textbf{x}_i,Z,\textbf{e}_i) = \{\rho_i(\textbf{x}_i,Z,\textbf{e}_i), \alpha_i(\textbf{x}_i,Z,\textbf{e}_i),\beta_i(\textbf{x}_i,Z,\textbf{e}_i)\}$.

Weights are optimised using Adam \citep{kingma2014adam}, with the learning rate set to $5\times10^{-4}$. Each model is trained for 100 epochs on 456 batches of 16 days each, using early stopping with a patience of 10 epochs.

\subsection{Experiments and datasets}
Having addressed the first aim in developing the convCNP model, we next evaluate model performance via three experiments. The first experiment compares the convCNP model to an ensemble of existing downscaling methods following a standardised experimental protocol. In contrast to the convCNP model, these methods are unable to make predictions at locations where training data are not available. In the second experiment, we assess the performance of the convCNP model at these unseen locations compared to a baseline constructed by interpolating single-site models. Finally, ablation experiments are performed to quantify the impact of including sub-grid scale topographic information on performance. 

\subsubsection{Experiment 1 - baseline comparison}
ConvCNP model performance is first compared to strong baseline methods taken from the VALUE experimental protocol. VALUE \citep{maraun2015value} provides a standardised suite of experiments to evaluate new downscaling methods, together with data benchmarking the performance of existing methods. In the VALUE 1a experiment, each downscaling method predicts the maximum temperature and daily precipitation at 86 stations across Europe (Figure 2), given gridded data from the ERA-Interim reanalysis \citep{dee2011era}. These stations are chosen as they offer continuous, high fidelity data over the training and held out test periods and represent multiple different climate regimes \citep{gutierrez2019intercomparison}. Data is taken from 1979-2008, with five-fold cross validation used over six-year intervals to produce a 30 year time-series. 

\begin{figure}
\centering
\includegraphics[scale=0.125]{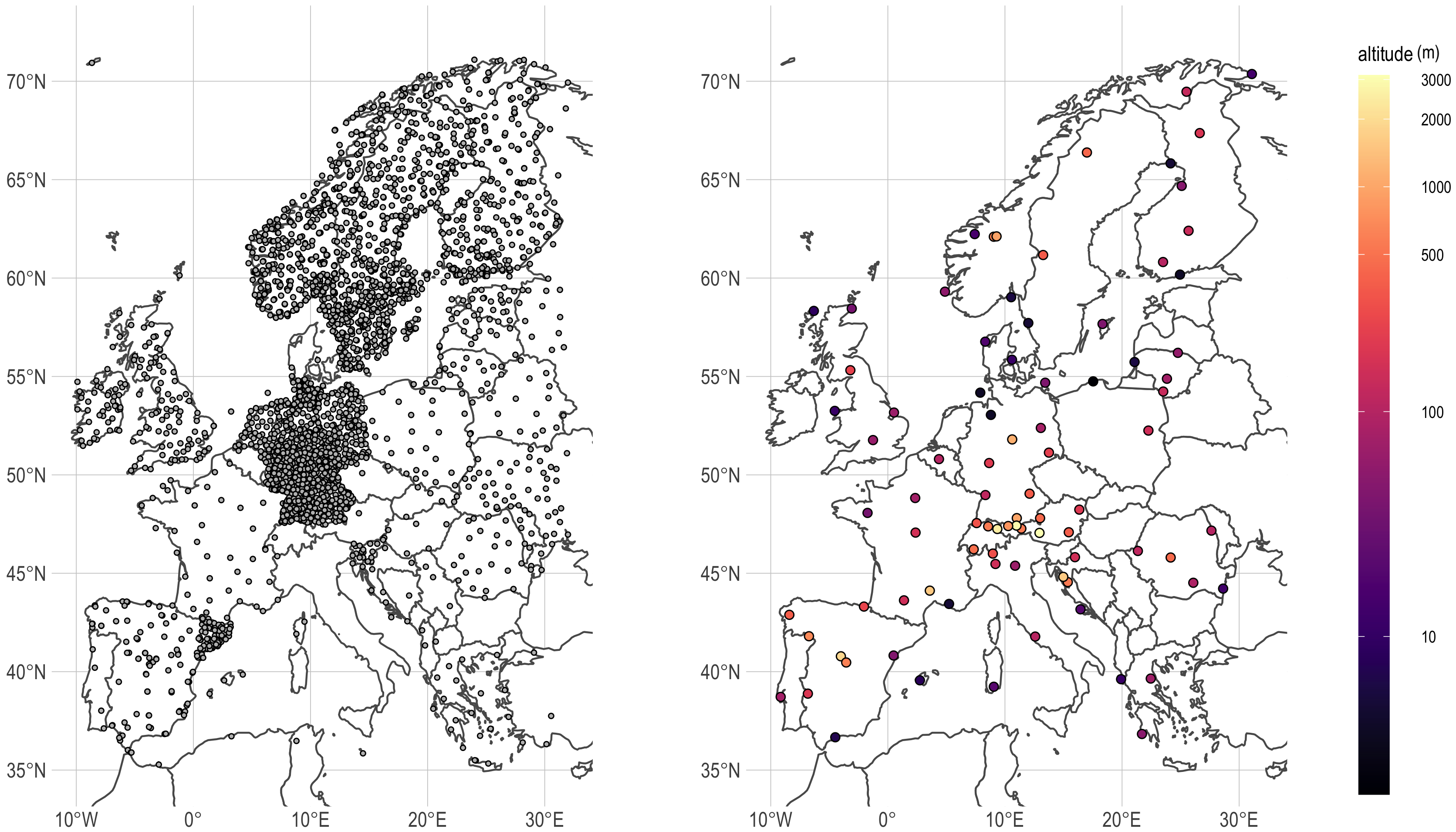}
\caption{Locations of ECA\&D (left) and VALUE (right) stations, with altitude shaded.}	
\end{figure}

The convCNPs are trained to predict maximum temperature and precipitation at these 86 VALUE stations given the ERA Interim grids over Europe. Station data are taken from the European Climate Assessment Dataset \citep{klein2002daily}. These grids are restricted to points between 35 and 72 degrees latitude and -15 to 40 degrees longitude. The VALUE experiment protocol does not specify which predictors are used in the downscaling model (i.e which gridded variables are included in $Z$). Based on the predictors used by methods in the baseline ensemble, winds, humidity and temperature are included at multiple levels together with time, latitude, longitude and invariant fields. Predictors are summarised in Table 1.

\begin{table}[t]
\begin{center}
\begin{tabular}{ccl}
\hline
 Predictor& Level & Description\\
 \hline
   \multicolumn{3}{c}{Surface} \\
 \hline
  TMAX & surface & Maximum temperature\\
  TMEAN & surface & Mean temperature\\
  U10 &surface& Northward wind\\
  V10 &surface& Eastward wind\\
  Pr & surface& Accumulated precipitation\\
  \hline
  \multicolumn{3}{c}{Upper level} \\
 \hline
  Q&850hPa, 700hPa, 500hPa& Specific humidity\\
  TA&850hPa, 700hPa, 500hPa& Temperature\\
  UA&850hPa, 700hPa, 500hPa& Northward wind\\
  VA& 850hPa, 700hPa, 500hPa& Eastward wind\\
  \hline
  \multicolumn{3}{c}{Invariant} \\
  \hline
  ASO&surface &Angle of sub-gridscale orography\\
  ANSO& surface & Anisotropy of sub-grid scale orography\\
  FSO &surface& Standard deviation of filtered subgrid orography\\
  SDO & surface & Standard deviation of orography\\
  GSFC& surface& Geopotential\\
  LAT& surface& Latitude\\
  LON& surface& Longitude\\
  \hline
  \multicolumn{3}{c}{Temporal} \\
  \hline
  Time& - &Day of year, transformed as $(cos(time), sin(time))$\\
  \hline
\end{tabular}
\end{center}
\caption{Gridded predictors from ERA-Interim reanalysis included in $Z$.}\label{t1}
\end{table}

For the sub-grid scale information for input into the final MLP, the point measurement of three products is provided at each station. True station elevation is taken from the Global Multi-resolution Terrain Elevation Dataset \citep{gmtr}. This is provided to the model together with the difference between the ERA-Interim gridscale resolution elevation and true elevation. Finally, topographic prominence is quantified using the ALOS Global mTPI \citep{theobald2015ecologically}. 

Results of the convCNP model are compared to all available PP models in the VALUE ensemble, a total of 16 statistical models for precipitation and 23 for maximum temperature. These models comprise a range of techniques including analogs, multiple linear regression, generalised multiple linear regression and genetic programming. For a complete description of all models included in the comparison, see Appendix A. 

\subsubsection{Experiment 2 - performance at unseen locations}
We next quantify model performance at unseen locations compared to an interpolation baseline. The convCNP models are retrained using station data from the European Climate Assessment Dataset (ECA\&D), comprising 3010 stations for precipitation and 3047 stations for maximum temperature (Figure 2). The 86 VALUE stations are held out as the validation set, testing the model performance at both unseen times and locations. 

As existing downscaling models are unable to handle unseen locations, it is necessary to construct a new baseline. A natural baseline for this problem is to construct individual models for each station using the training set, use these to make predictions at future times, and then interpolate to get predictions at the held out locations. For the single-station models, predictors are taken from ERA-Interim data at the closest gridbox, similar to \citet{gutierrez2013reassessing}. Multiple linear regression is used for maximum temperature. For precipitation, occurrence is modelled using logistic regression, and accumulation using a generalised linear model with gamma error distribution, similar to \citet{san2017reassessing}. These methods are chosen as they are amongst the best-performing methods of the VALUE ensemble for each variable \citep{gutierrez2019intercomparison}. 

Following techniques used to convert station observations to gridded datasets \citep{haylock2008european}, predictions at these known stations in the future time period are made by first interpolating monthly means (totals) for temperature (precipitation) using a thin-plate spline, then using a GP to interpolate the anomalies (fraction of the total value). All interpolation is three dimensional over longitude, latitude and elevation. Throughout the results section, this model is referred to as the GP-baseline.

\subsubsection{Experiment 3 - topography ablation}
Finally, the impact of topography on predictions is quantified. Experiment 2 is repeated three times with different combinations of topographic data fed into the final MLP (step 3 in Figure 1): no topographic data, elevation and elevation difference only and mTPI only. 

\subsection{Evaluation metrics}
A selection of standard climate metrics are chosen to assess model performance over the evaluation period, quantifying the representation of mean properties and extreme events (Table 2). Metrics are chosen based on those reported for the VALUE baseline ensemble \citep{gutierrez2019intercomparison,widmann2019validation,maraun2019value,hertig2019comparison}. 

Comparison to these metrics requires generating a timeseries of values from the distributions predicted by the convCNP model. For temperature, this is generated by taking the mean of the predicted distribution for mean metrics, and sampling is used to complete the extreme metrics. For precipitation, a day is first classified as wet if $\rho \geq 0.5$ or dry if $\rho < 0.5$. For wet days,  accumulations are generated by taking the mean of the gamma distribution for mean metrics or sampling for extreme metrics. 

\begin{table}[t]
\begin{center}
\begin{tabular}{p{2.2cm}p{2.2cm}p{9cm}}
\hline
 \multicolumn{3}{c}{Means} \\
 \hline
 Metric& Variables & Description\\
 \hline
  mb & $T_{max}$, Precip & mean bias\\
  sp & $T_{max}$, Precip & Spearman correlation between observed and predicted timeseries\\
  MAE &$T_{max}$, Precip& Mean absolute error\\
  R01 &Precip& Relative wet day frequency (predicted precipitation days: observed precipitation days. \\
  SDII & Precip& Mean wet day precipitation\\
 \hline 
 \multicolumn{3}{c}{Extremes} \\
 \hline
 Metric& Variables & Description\\
 \hline
  98P & $T_{max}$, precip & Bias in the 98th percentile.\\
  R10 & Precip & Relative frequency of days with precipitation greater than 10mm\\
  \hline 
\end{tabular}
\end{center}
\caption{Evaluation metrics.}
\end{table}
%%%%%%%%%%%%%%%%%%%%%
% BASELINE COMPARISON
%%%%%%%%%%%%%%%%%%%%%

\section{Results: baseline comparison (experiment 1)}
The convCNP model outperforms all VALUE baselines on median mean absolute error (MAE) and spearman correlation for both maximum temperature and precipitation. Comparisons of convCNP model performance at the 86 VALUE stations to each model in the VALUE baseline ensemble are shown in Figure 4. The low MAE and high spearman correlation indicate that the model performs well at capturing day-to-day variability. 

\begin{figure}
\centering
\includegraphics[scale=0.14]{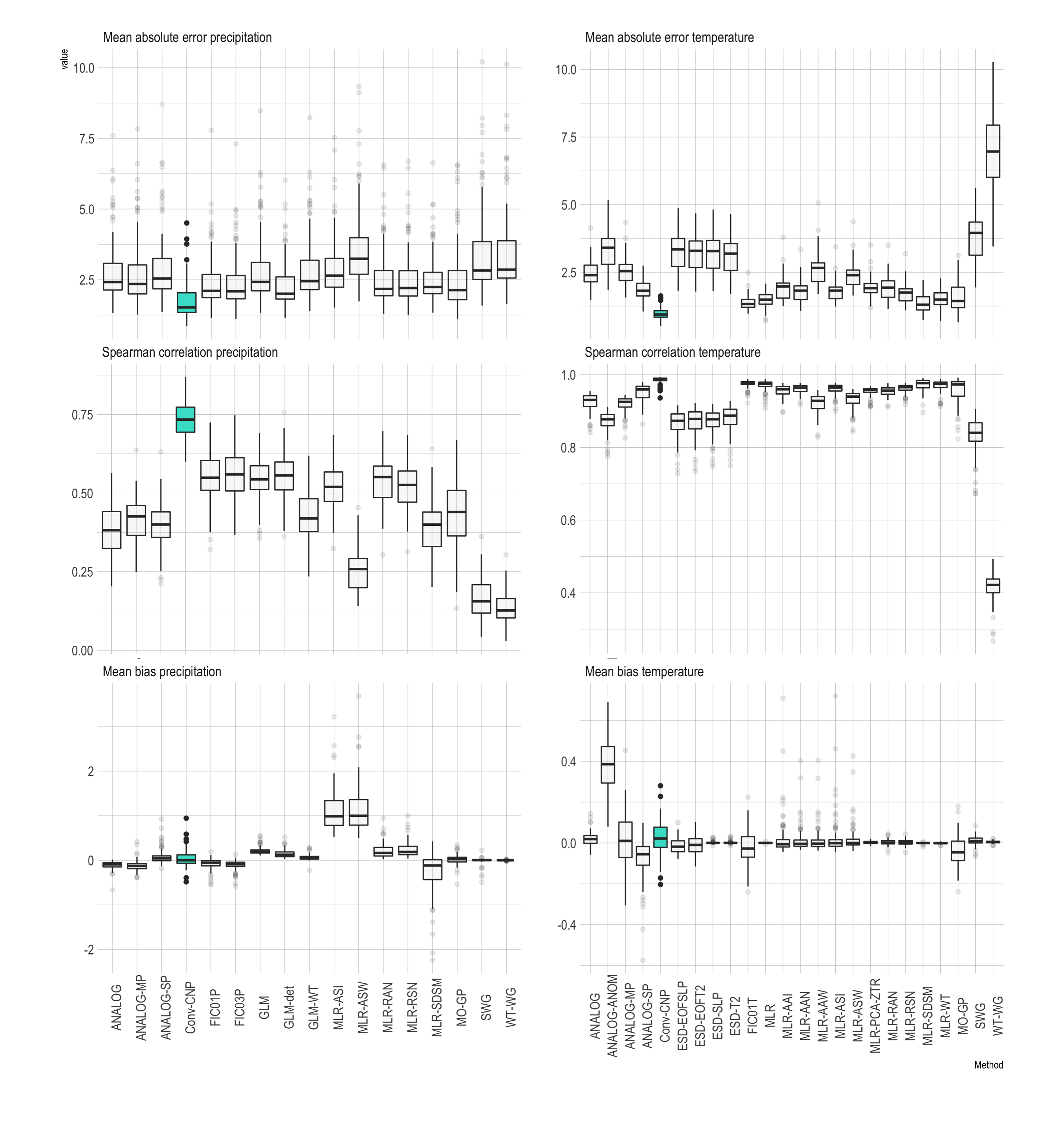}
\caption{Comparison of the convCNP model to VALUE ensemble baselines for mean metrics, with the convCNP model shaded in blue. Each box summarises performance for one model in the ensemble over the 86 training stations on the held out validation data.}	
\end{figure}

For maximum temperature, the mean bias is larger than baseline models at many stations, with interquartile range -0.02C to 0.08C. This is a direct consequence of training a global model as opposed to individual models to each station which will trivially correct the mean \citep{maraun2018statistical}. Though larger than baseline models, this error is still small for a majority of stations. Similarly for precipitation, though mean biases are larger than many of the VALUE models, the interquartile range is just -0.07mm to 0.12mm. For precipitation, the bias in convCNP relative wet day frequency (R01) and mean wet day precipitation (SDII) are comparable to the best models in the VALUE ensemble (not shown). 

When downscaling GCM output for impact studies, it is of particular importance to accurately reproduce extreme events \citep{katz1992extreme}. In line with previous work comparing the VALUE baselines \citep{hertig2019comparison}, an extreme event is defined to be a value greater than the 98th percentile of observations. Comparisons of biases in the 98th percentile of maximum temperature and precipitation are shown in Figure 5. The convCNP performs similarly to the best baselines, with a median bias of -0.02C for temperature and -2.04mm for precipitation across the VALUE stations. R10 biases are comparable to baselines, with a median bias of just -0.003mm. 

\begin{figure}
\centering
\includegraphics[scale=0.13]{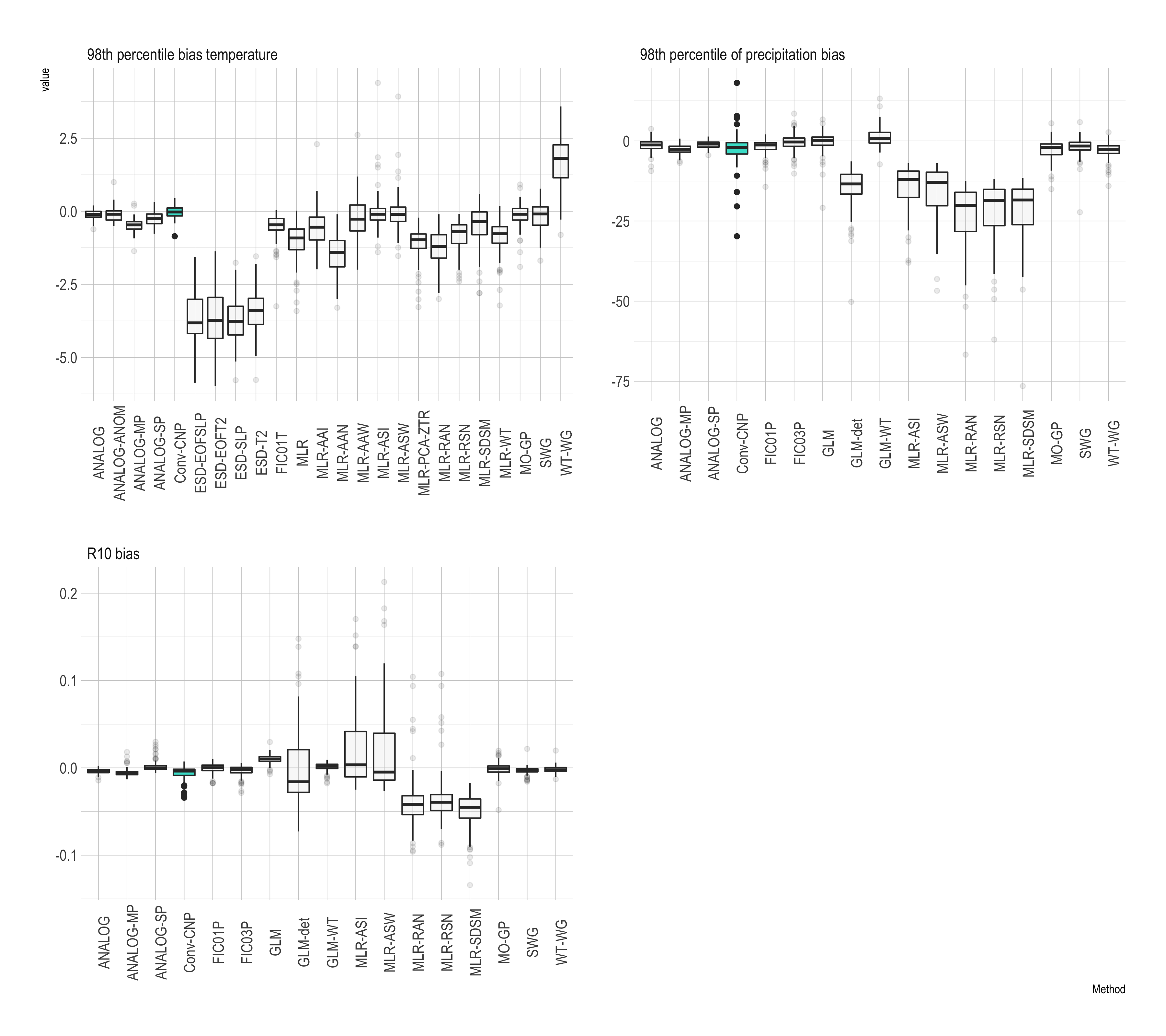}
\caption{As for Figure 4, but for extreme metrics. }	
\end{figure}

%%%%%%%%%%%%%%%%%%%%%
% GENERALISATION
%%%%%%%%%%%%%%%%%%%%%
\section{Results: performance at unseen locations (experiment 2)}
The convCNP model outperforms the GP-baseline at unseen stations. Results for MAE, spearman correlation and mean bias are shown in Figure 6. For maximum temperature, the convCNP model gives small improvements over the baseline model, with spearman correlations of 0.99 (0.98) and MAE of 1.19C (1.35C) for the convCNP (GP baseline). Importantly, large outliers ($>$10C) in the baseline MAE are not observed in the convCNP predictions. Figure 7 shows the spatial distribution of MAE for the convCNP and GP-baseline together with the difference in MAE between the two models. This demonstrates that stations with high MAE in the convCNP model are primarily concentrated in the complex topography of the European Alps. The GP-baseline model displays large MAE not only in the Alps, but also at other locations for example in Spain and France. The convCNP improves predictions at 82 out of the 86 stations. 

\begin{figure}
\centering
\includegraphics[scale=0.125]{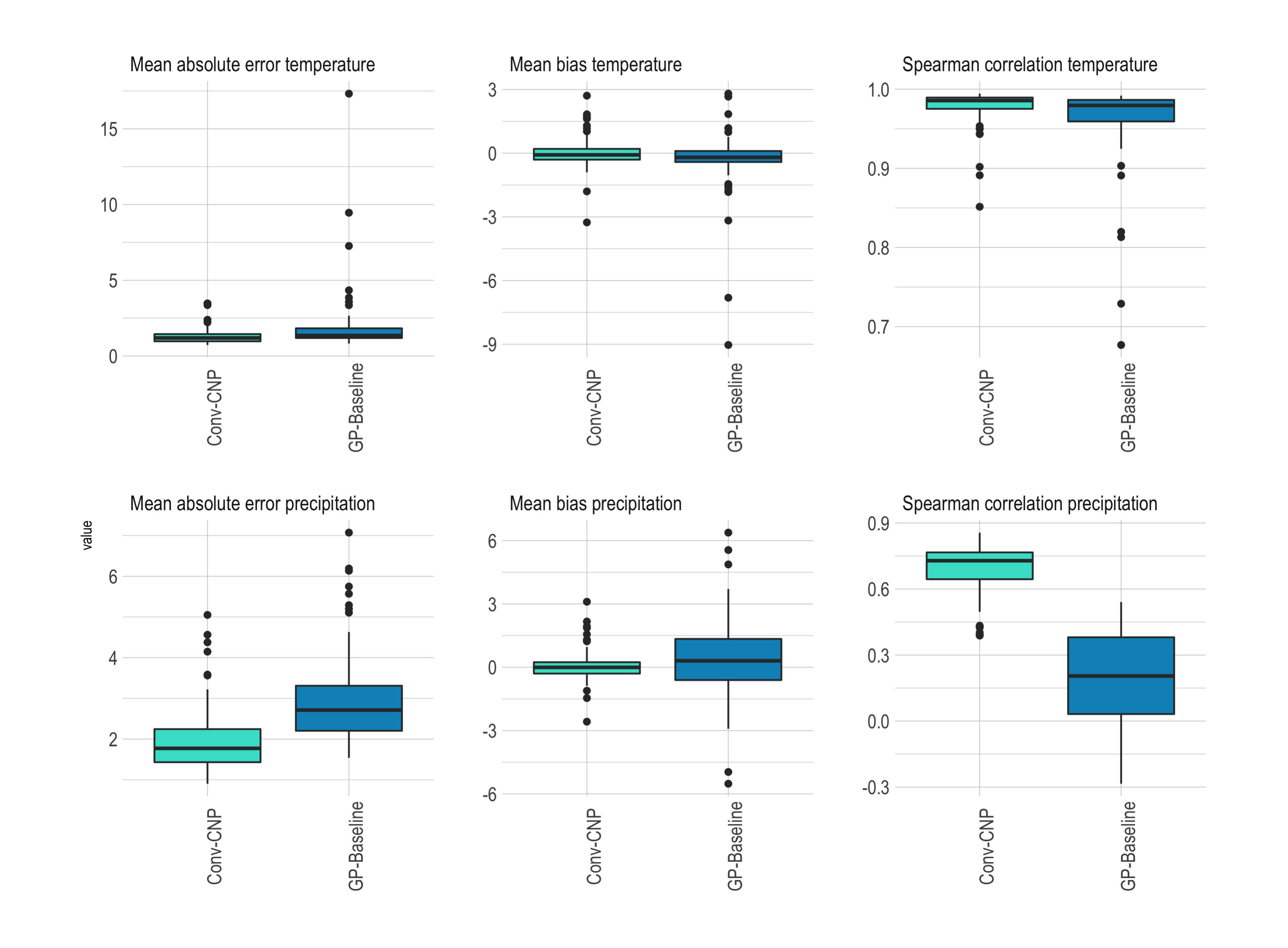}
\caption{Comparison of the convCNP model to the GP-baseline. Boxes summarise model performance over the 86 held out VALUE stations for maximum temperature (top) and precipitation (bottom) for each of the mean metrics.}	
\end{figure}

\begin{figure}
\begin{center}
\includegraphics[scale=0.125]{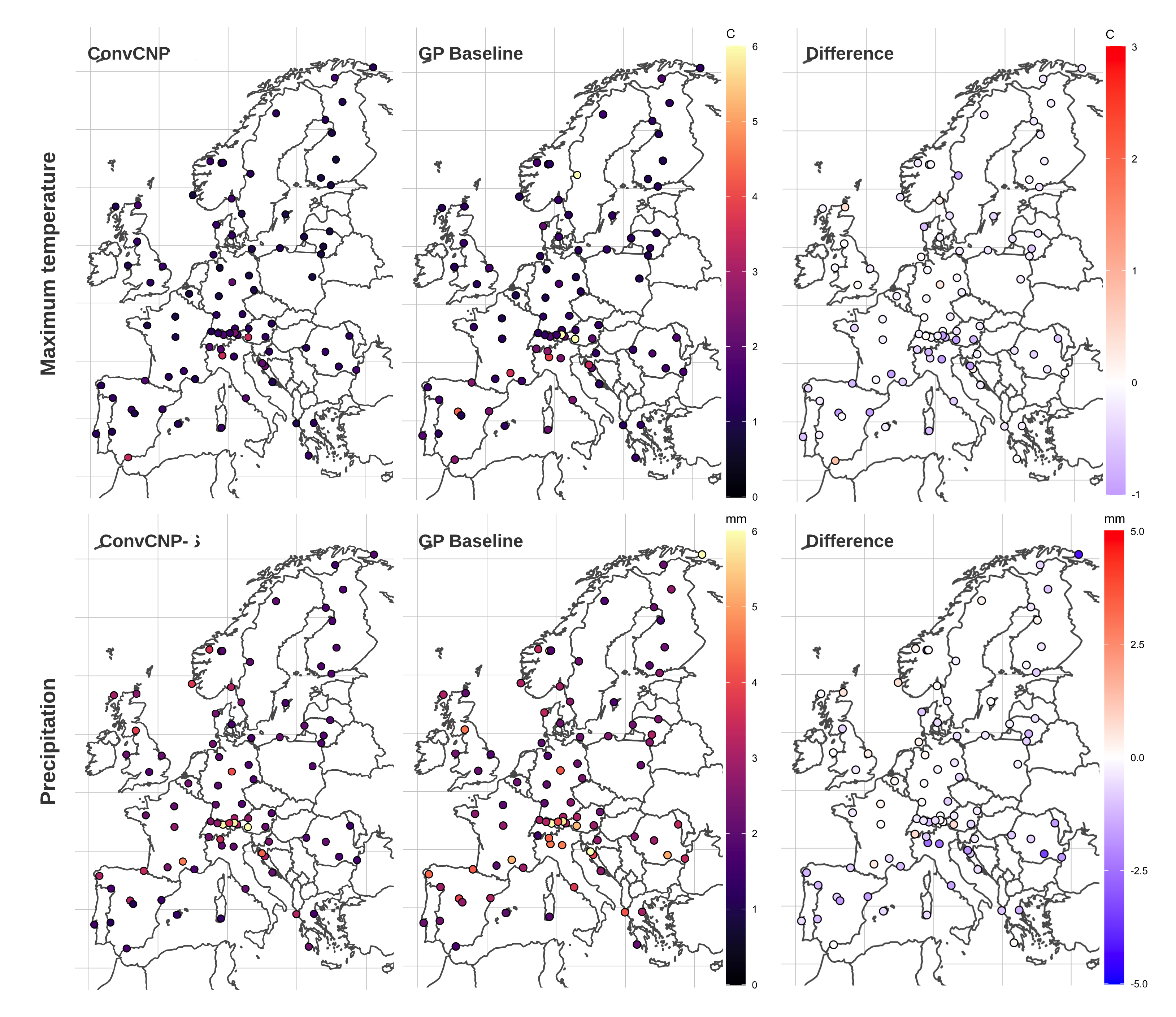}
\caption{Spatial distribution of mean absolute error for convCNP (right), GP baseline (centre) and convCNP - GP baseline (left). Maximum temperature (precipitation) is shown on the top (bottom) row. All panels show results for each of the 86 held out VALUE stations. }	
\end{center}
\end{figure}

Repeating this analysis for precipitation, the convCNP model gives substantial improvement over the baseline for MAE and spearman correlation. Spearman correlations are 0.57 (0.20) and MAE 2.10mm (2.71mm) for convCNP (GP-baseline). In contrast to maximum temperature, there is no clear link between topography and MAE, though again convCNP predictions have large MAE for multiple stations located in the Alps. The convCNP model improves on baseline predictions at 80 out of 86 stations. 

\begin{figure}
\begin{center}
\includegraphics[scale=0.12]{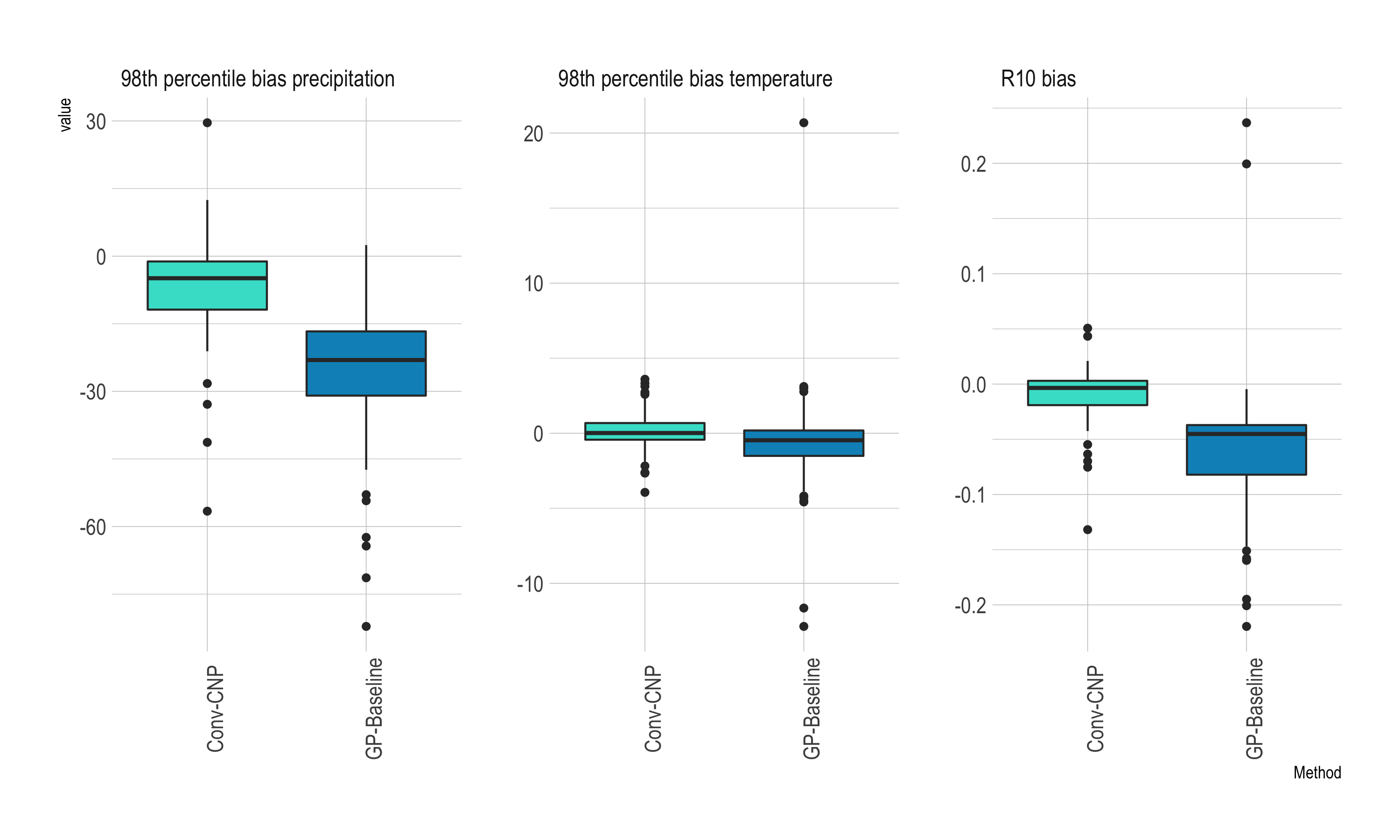}
\caption{As for Figure 6, but for P98 and R10 biases. }	
\end{center}
\end{figure}

Comparisons between models for extreme metrics are shown in Figure 8. For maximum temperature, the convCNP has slightly lower absolute 98th percentile bias than the baseline. For precipitation, errors are substantially lower, with median absolute 98th percentile bias of 4.90mm for convCNP compared to 22.92mm for the GP-baseline. The spatial distributions of 98th percentile bias for maximum temperature and precipitation predictions together with the difference in absolute bias are shown in Figure 9. For maximum temperature, the convCNP does not improve on the baseline at all stations. The GP-baseline exhibits uniformly positive biases, while the convCNP model has both positive and negative biases. Improvements are seen through central and eastern Europe, while the convCNP performs comparatively poorly in Southern Europe and the British Isles. For precipitation, predictions have low biases across much of Europe for the convCNP, with the exception of in the complex terrain of the Alps. GP-baseline biases are negative throughout the domain. For this case, convCNP predictions have lower bias at 84 of the 86 validation stations. 

\begin{figure}
\begin{center}
\includegraphics[scale=0.12]{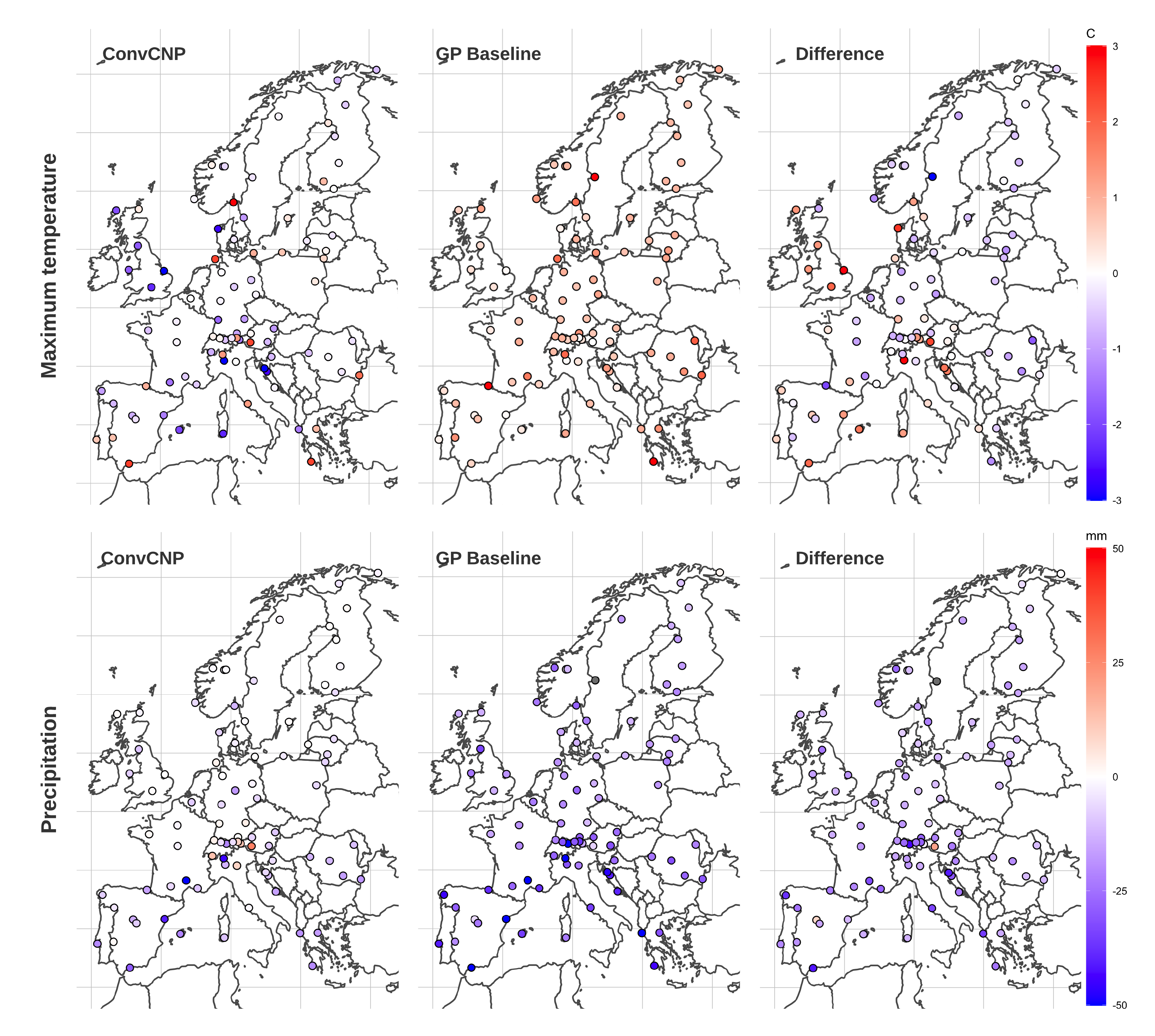}
\caption{As for Figure 7, but for P98 biases. Here, the difference panels quantify the difference in absolute bias $|P98_{convCNP}| - |P98_{GP-Baseline}|$. Negative (positive) values indicate that the convCNP (GP-Baseline) has better performance.}	
\end{center}
\end{figure}

A limitation to the analysis of the standard climate metrics in Table 2 is that these only assess certain aspects of the predicted distribution. To assess the calibration of the models, we next examine the probability integral transform (PIT) values. The PIT value for a given prediction is defined as the CDF of the distribution predicted by the convCNP model evaluated at the true observed value. These values can be used to determine whether the model is calibrated by evaluating the PIT for every model prediction at the true observation, and plotting their distribution. If the model is properly calibrated, it is both necessary and sufficient for this distribution to be uniform \citep{gneiting2007probabilistic}. PIT distributions for maximum temperature and wet-day precipitation are shown in Figure 10. For temperature, the model is well calibrated overall, although the predicted distributions are often too narrow, as demonstrated by the peaks around zero and one indicating that the observed value falls outside the predicted normal distribution. Calibration of the precipitation model is poorer overall. The peak in PIT mass around zero indicates that this model often over predicts rainfall accumulation. Performance varies between individual stations for both temperature and precipitation, with examples of PIT distributions for both well- and poorly-calibrated stations shown in Figure 10.

\begin{figure}
\begin{center}
\includegraphics[scale=0.125]{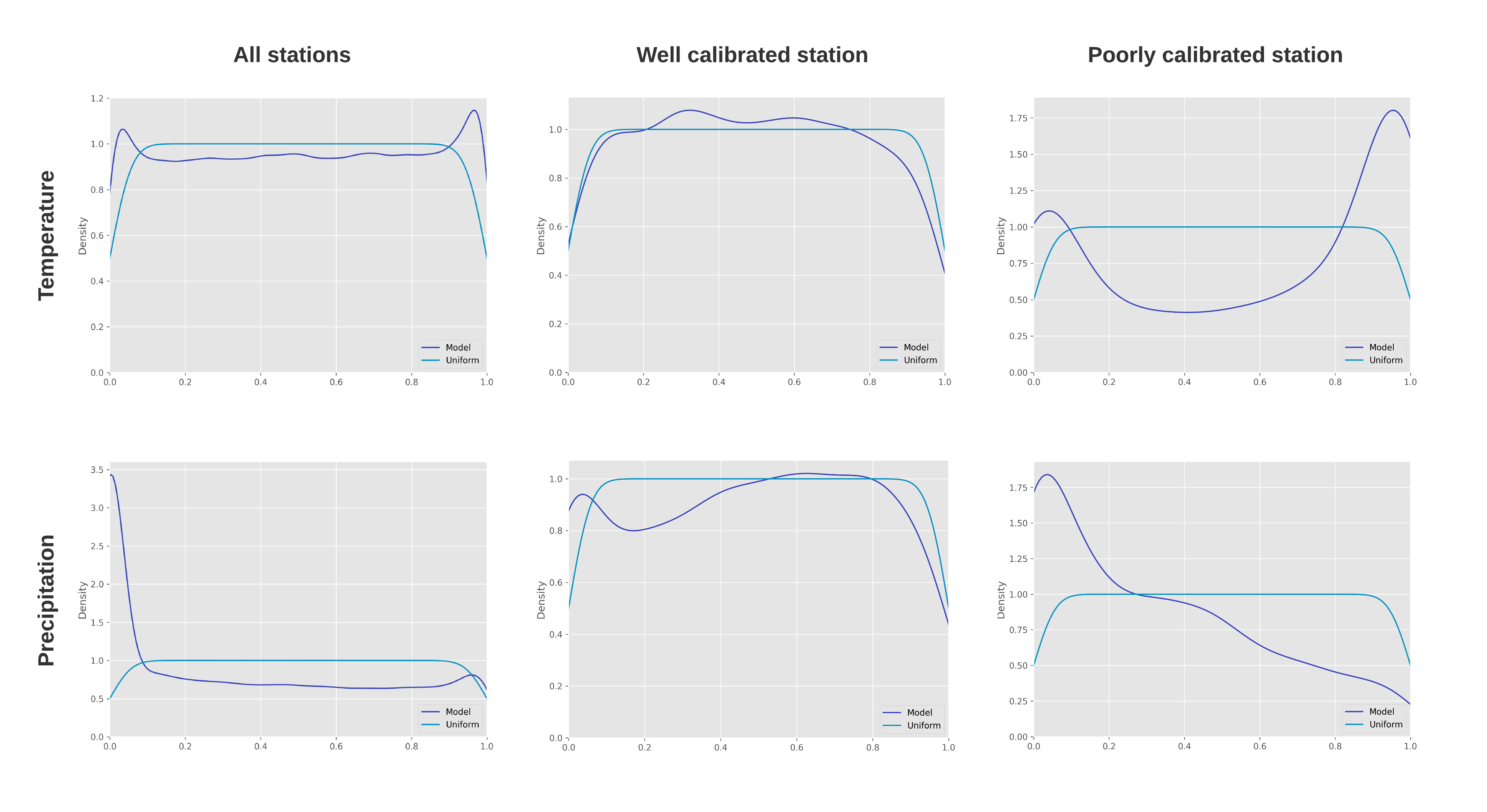}
\caption{KDEs showing probability integral transforms for model predictions compared to a uniform distribution to assess calibration for temperature (top) and precipitation (bottom). For temperature PIT plots are shown for all values (left), a station where the model is well calibrated (Braganca, Portugal; centre) and a station where the model is poorly calibrated (Gospic, Croatia; right). Similarly for precipitation, PIT plots are shown for all values (left), a station where the model is well calibrated (Stornoway, UK; centre) and a station where the model is poorly calibrated (Sondankyla, Finland; right). }	
\end{center}
\end{figure}

%%%%%%%%%%%%%%%%%%%%%
% TOPOGRAPHY
%%%%%%%%%%%%%%%%%%%%%
\section{Results: topography ablation (experiment 3)}
Results of the topography ablation experiment are shown in Figure 11 (mean metrics) and Figure 12 (extreme metrics). These figures compare the performance on each metric between the convCNP model with all topographic predictors, and convCNP models trained with no topography, elevation and elevation difference only and mTPI only. 

For maximum temperature, inclusion of topographic information improves MAE, mean bias and spearman correlation. Models including only mTPI or no topographic predictors have a number of stations with very large MAE, exceeding 10C at several stations. Unsurprisingly, these stations are found to be located in areas of complex topography in the Alps (not shown). Including elevation both decreases the median MAE and corrects errors at these outliers, with further improvement observed with mTPI added. A similar pattern is seen for mean bias. More modest improvements are seen for precipitation, though inclusion of topographic data does result in slightly improved performance. 

\begin{figure}
\begin{center}
\includegraphics[scale=0.12]{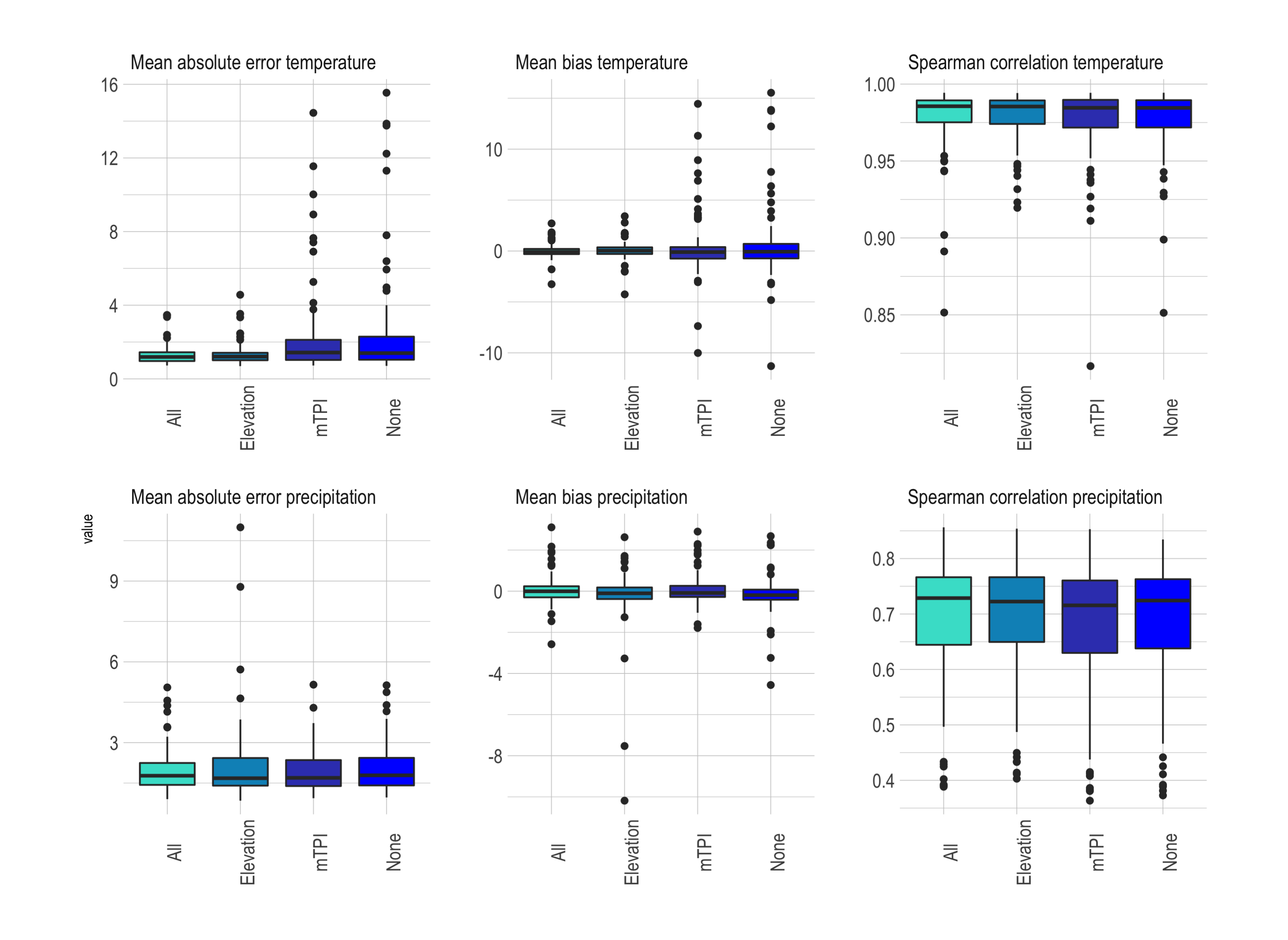}
\caption{Comparison of model performance in the topography ablation experiment. The complete model (All) is compared to models with no topographic data (None), elevation and elevation difference only (Elevation) and mTPI only (mTPI). Boxes summarise model performance over the 86 held out VALUE stations for maximum temperature (top) and precipitation (bottom) for each of the mean metrics.}	
\end{center}
\end{figure}

For maximum temperature, inclusion of topographic data results in reduced 98th percentile bias. This is primarily as a result of including elevation and elevation difference data, with limited benefit derived from the inclusion of mTPI. In contrast, for precipitation, models with topographic correction perform worse than the elevation agnostic model for both 98th percentile and R10 biases. This reduced performance for precipitation may result from overfitting.

\begin{figure}
\begin{center}
\includegraphics[scale=0.12]{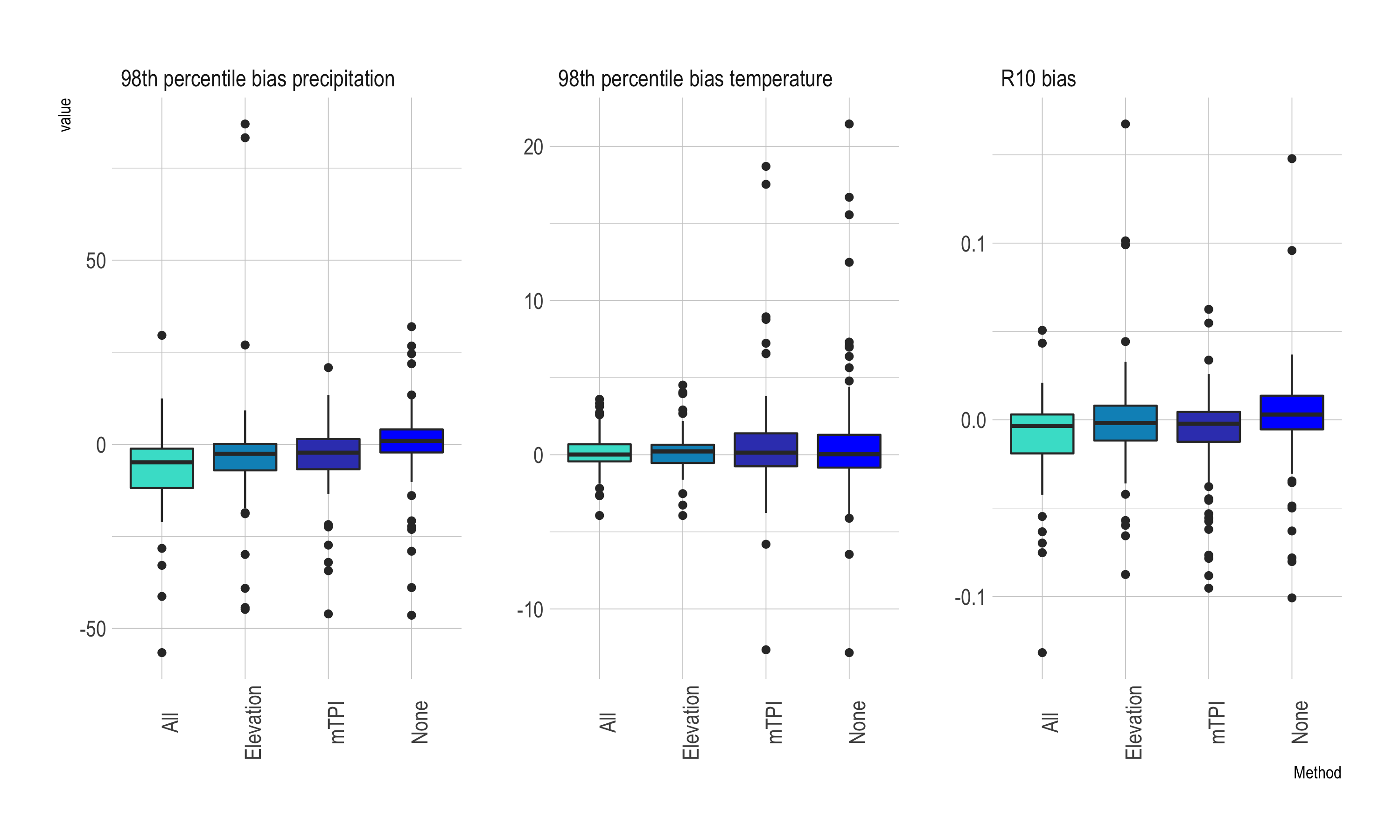}
\caption{As for Figure 11, but for extreme metrics. }	
\end{center}
\end{figure}

%%%%%%%%%%%%%%%%%%%%%
% CONCLUSION
%%%%%%%%%%%%%%%%%%%%%
\section{Discussion and conclusion}

This study demonstrated the successful application on convCNPs to statistical downscaling of temperature and precipitation. The convCNP model performs well compared to strong baselines from the VALUE ensemble on both mean and extreme metrics. For both variables the convCNP model outperforms an interpolation based baseline. Inclusion of sub-grid-scale topographic information is shown to improve model performance for mean and extreme metrics for maximum temperature, and mean metrics for precipitation. 

The convCNP model has several advantages over existing methods beyond performance on these metrics. The most important of these is that the trained model can be used to predict values at any target point, regardless of whether this point is included in the training set. In Sect. 2, we demonstrated that the convCNP model outperforms interpolation of single-site downscaling models. This has application in developing local projections for climate impact studies. A second useful feature is that the convCNP takes raw grid data as input, without the need for feature engineering. This contrasts with many of the baseline methods, which rely on the identification of relevant circulation analogs. Although only temperature and precipitation are considered in this study, the model is easily applied to any climate variable with available station observations, for example windspeed. 

It is instructive to consider these results more broadly in the context of the debate surrounding the application of deep learning models in downscaling. In this study we deploy deep learning approaches using domain-specific knowledge to shape the architecture, and then deploy this inside simple statistical observation models. This approach is demonstrated to improve on simple baselines on both mean and extreme metrics for maximum temperature and precipitation. This contrasts to previous work suggesting that little benefit is derived from applying neural network models to downscaling maximum temperature \citep{bano2020configuration} and precipitation \citep{vandal2019intercomparison}, and motivates continued efforts to develop and benchmark such models for application in climate impact studies. 

Several areas remain for future work. Representation of certain metrics, notably precipitation extremes requires further improvement, particularly in areas with complex topography. The topography ablation experiments demonstrate that the convCNP P98 bias increases with the inclusion of topographic information. Dynamically, this is likely due to local flow effects such as F\"{o}hn winds \citep{gaffin2007foehn,basist1994statistical}, which depend on the incident angle of the background flow. A possible explanation for this is that the MLP is insufficient to model these effects. Further experimentation with adding a second CNN to capture the sub-grid-scale processes, and possibly conditioning predictions of this model on local flow is left as a topic for future research. 

Another avenue for improving model performance would be to change the distribution predicted by the convCNP. Model calibration results presented in Section 4 indicate that the temperature downscaling model could be improved using a distribution with heavier tails. Precipitation model calibration requires improvement, with the model frequently under-predicting wet day accumulations. A possible explanation for this is that the left hand tail of the gamma distribution decays rapidly. For cases where the mode of the predicted distribution is greater than zero, small observed accumulations are therefore heavily penalised. Previous work has acknowledged that the Bernoulli-Gamma distribution used in this study is not realistic for all sites \citep{vlvcek2009daily}, and suggested that representation of precipitation extremes can be improved using a Bernoulli-Gamma-Gerneralised Pareto distribution \citep{ben2015probabilistic,volosciuk2017combined}. Future work will explore improving the calibration of the downscaling models using mixture distributions and normalising flows \citep{rezende2015variational} to improve the calibration of the model. A further possibility for extending the convCNP model would be to explicitly incorporate time by building recurrence into the model \citep{qin2019recurrent,singh2019sequential}.

The final aspect to consider is extending these promising results downscaling reanalysis data to apply to future climate simulations from GCMs. An in depth analysis of the convCNP model performance on seasonal and annual metrics would be beneficial in informing application to impact scenarios. A limitation in all PP downscaling techniques is that applying a transfer function trained on reanalysis data to a GCM makes the assumption that the predictors included in the context set are realistically simulated in the GCM \citep{maraun2018statistical}. Future work will aim to address this issue through training a convCNP model directly on RCM or GCM hindcasts available through projects such as EURO-CORDEX \citep{jacob2014euro}.
 
 \codeavailability{Model code is available at https://github.com/anna-184702/convCNPClimate.} %% use this section when having only software code available
 
 \appendix
 \section{VALUE ensemble methods}
Table 3 summarises the baseline methods in the VALUE ensemble. This information is adapted from \citep{gutierrez2019intercomparison}. 

\newpage
\begin{table}[t]
\begin{center}
\begin{tabular}{p{2.0cm}p{2.0cm}p{13cm}}
\hline
\centering
 Model& Variables & Description\\
 \hline
  ANALOG & $T_{max}$, precip & Standard analog, no seasonal component \citep{gutierrez2013reassessing}\\
  ANALOG-ANOM & $T_{max}$ & Analog with seasonal component \citep{ayar2016intercomparison}\\
  ANALOG-MP &$T_{max}$, precip& Analog method with seasonal component \citep{raynaud2017atmospheric}\\
  ANALOG-SP &$T_{max}$, precip&  Analog method with seasonal component \citep{raynaud2017atmospheric} \\
  ESD-EOFSLP & $T_{max}$&Multiple linear regression \citep{benestad2015using} \\
  ESD-SLP & $T_{max}$& Multiple linear regression \citep{benestad2015using} \\
  ESD-T2 & $T_{max}$& Multiple linear regression \citep{benestad2015using}\\
  FIC01P & $T_{max}$, precip& Two step analog method \citep{ribalaygua2013description} \\
  FIC03P & precip& Two step analog method \citep{ribalaygua2013description} \\
  GLM & precip& Generalised linear Model with log-canonical link function \citep{san2017reassessing}. Bernoulli error distribution for occurrence and gamma error distribution for accumulation. Predictions sampled from output distribution.   \\
  GLM-det & precip& As for GLM, predictions given as mean of output distribution \citep{san2017reassessing}.\\
  GLM-WT & precip& As for GLM, conditioned on 12 weather types identified by k-means clustering \citep{san2017reassessing}.  \\
  MLR & $T_{max}$& Multiple linear regression using PCA for predictors \citep{gutierrez2013reassessing}.\\
  MLR-AAI & $T_{max}$, precip& Multiple linear regression, annual training, anomaly data, inflation variance correction \citep{huth2015comparative}.\\
  MLR-AAN & $T_{max}$, precip& Multiple linear regression, annual training, anomaly data, white-noise variance correction \citep{huth2015comparative}. \\
  MLR-AAW & $T_{max}$, precip& Multiple linear regression, annual training, anomaly data, white-noise variance correction \citep{huth2015comparative}.\\
  MLR-ASI & $T_{max}$, precip& Multiple linear regression, seasonal training, anomaly data, inflation variance correction \citep{huth2015comparative}.\\
  MLR-ASW & $T_{max}$, precip& Multiple linear regression, seasonal training, anomaly data, white-noise variance correction \citep{huth2015comparative}.\\
  MLR-PCA-ZTR & $T_{max}$& Multiple linear regression with s-mode principal component predictors \citep{jacobeit2014statistical}.\\
  MLR-RAN & $T_{max}$, precip& Multiple linear regression, seasonal training, raw data, no variance correction \citep{huth2015comparative}.\\
  MLR-RSN & $T_{max}$, precip& Multiple linear regression, seasonal training, raw data, no variance correction \citep{huth2015comparative}.\\
  MLR-SDSM & $T_{max}$, precip& Single-site multiple linear regression using the statistical downscaling method \citep{wilby2002sdsm}.\\
  MLR-WT & $T_{max}$& As for MLR, but conditioned on weather types defined using k-means clustering \citep{gutierrez2013reassessing}.\\
  MO-GP & $T_{max}$, precip& Multi-objective genetic programming \citep{zerenner2016downscaling}. \\
  SWG & $T_{max}$, precip& Two-step vectorised generalised linear models \citep{ayar2016intercomparison}.\\
  WT-WG & $T_{max}$, precip& Distributional fitting based on weather types selected using k-means clustering \citep{gutierrez2013reassessing}.\\
\hline
\end{tabular}
\caption{Summary of models included in the VALUE ensemble.}\label{t1}
\end{center}
\end{table}

\authorcontribution{
AV implemented the code, conducted the experiments and wrote the first draft. All authors designed the study and contributed to the analysis of results and final version of the paper. 
} %% this section is mandatory

\competinginterests{The authors declare that they have no conflict of interest.} %% this section is mandatory even if you declare that no competing interests are present

\begin{acknowledgements}
Anna Vaughan acknowledges the UKRI Centre for Doctoral Training in the Application of Artificial Intelligence to the study of Environmental Risks (AI4ER), led by the University of Cambridge and British Antarctic Survey, and studentship funding from Google DeepMind. Will Tebbutt is supported by Google DeepMind.
\end{acknowledgements}

%% REFERENCES

%% The reference list is compiled as follows:

%% Since the Copernicus LaTeX package includes the BibTeX style file copernicus.bst,
%% authors experienced with BibTeX only have to include the following two lines:
%%
\bibliographystyle{copernicus}
\bibliography{oceanbib}

\end{document}